\documentclass[journal,comsoc]{IEEEtran}
\usepackage[T1]{fontenc}

\usepackage{ifpdf}
\usepackage{cite}
\ifCLASSINFOpdf
  \usepackage[pdftex]{graphicx}
\else
  \usepackage[dvips]{graphicx}
\fi
\usepackage{amsmath}
\usepackage[cmintegrals]{newtxmath}
\usepackage{bm}
\usepackage{algorithmic}
\usepackage[ruled,linesnumbered]{algorithm2e}
\usepackage{array}
\ifCLASSOPTIONcompsoc
  \usepackage[caption=false,font=normalsize,labelfont=sf,textfont=sf]{subfig}
\else
  \usepackage[caption=false,font=footnotesize]{subfig}
\fi
\ifCLASSOPTIONcaptionsoff
  \usepackage[nomarkers]{endfloat}
 \let\MYoriglatexcaption\caption
 \renewcommand{\caption}[2][\relax]{\MYoriglatexcaption[#2]{#2}}
\fi
\usepackage{url}
\usepackage[backref, hidelinks]{hyperref}
\usepackage{bbm}
\usepackage[normalem]{ulem}
\def \j{\bm{\mathit{j}}}
\def \J{\bm{\mathit{J}}}
\def \I{\mathit{I}}
\def \H{\mathit{H}}
\begin{document}
\title{Context-Aware Deep Spatio-Temporal Network for Hand Pose Estimation from Depth Images}

\author{{Yiming Wu, Wei Ji, Xi Li*, Gang Wang, Jianwei Yin and Fei Wu}
\thanks{Y. Wu, W. Ji, J. Yin, F. Wu are with College of Computer Science, Zhejiang University, Hangzhou 310027, China (e-mail: ymw, jiwei@zju.edu.cn zjuyjw, wufei@cs.zju.edu.cn).}
\thanks{X. Li*(corresponding author) is with the College of Computer Science and Technology, Zhejiang University, Hangzhou 310027, China, and also with the Alibaba-Zhejiang University Joint Institute of Frontier Technologies, Hangzhou 310027, China (e-mail: xilizju@zju.edu.cn).}
\thanks{Gang Wang is with AI Labs of Alibaba Group, Hangzhou 310027, China (email: gangwang6@gmail.com).}}

\markboth{IEEE TRANSACTIONS ON CYBERNETICS}%
{Wu \MakeLowercase{\textit{et al.}}: Context-Aware Deep Spatio-Temporal Network for Hand Pose Estimation from Depth Images}

\maketitle

\begin{abstract}
As a fundamental and challenging problem in computer vision, hand pose estimation
aims to estimate the hand joint locations from depth images. Typically, the problem is modeled as
learning a mapping function from images to hand joint coordinates in a data-driven manner.
In this paper, we propose Context-Aware Deep Spatio-Temporal Network (CADSTN), a novel method to jointly
model the spatio-temporal properties for hand pose estimation. Our proposed network is able to learn the
representations of the spatial information
and the temporal structure from the image sequences. Moreover, by adopting adaptive fusion method,
the model is capable of dynamically weighting different predictions to lay emphasis on sufficient context.
Our method is examined on two common benchmarks,
the experimental results demonstrate that our proposed approach achieves
the best or the second-best performance with state-of-the-art methods and runs in 60fps.
\end{abstract}

\ifCLASSOPTIONpeerreview
\begin{center} \bfseries EDICS Category: 3-BBND \end{center}
\fi
%
\IEEEpeerreviewmaketitle

\section{Introduction}\label{sec:introduction}
\begin{figure*}[htbp]
    \centering
    \includegraphics[width=1\linewidth]{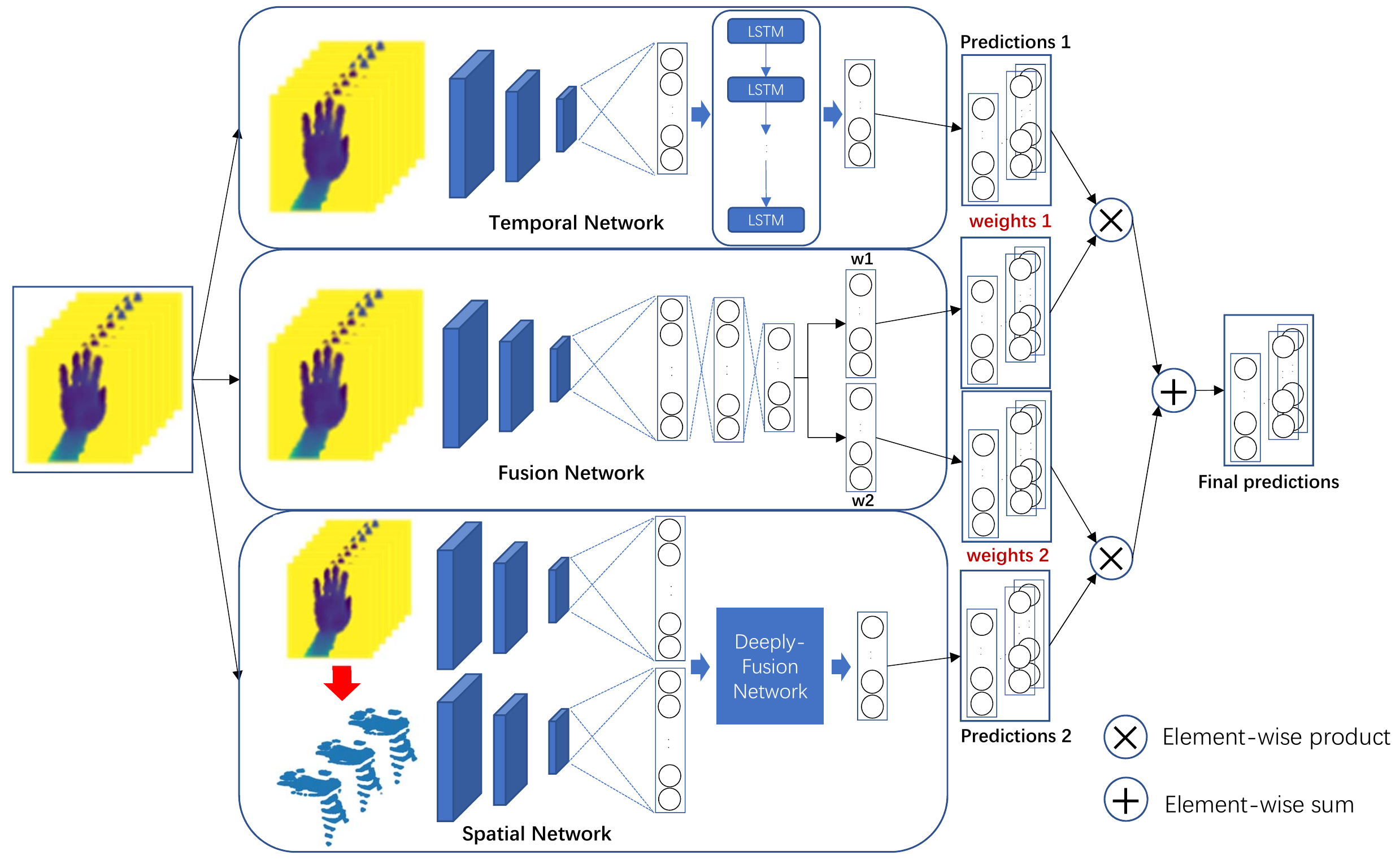}
    \caption{The overview of total network. The network is composed of three parts: \emph{Temporal Network},
    \emph{Spatial Network} and \emph{Fusion Network}.
    (top) \emph{Temporal Network} extracts the features considering the temporal coherence of input images via LSTM,
    and outputs a sequence of joint locations.
    (bottom) \emph{Spatial Network} employs the Deeply-Fusion
    framework to hierarchically fuse the features from a single depth image and corresponding sliced 3D volumetric representations,
    then gives out prediction.
    (center) \emph{Fusion Network} fuses the predictions from two networks, it learns the weights for each prediction
    and the final prediction is calculated as summation of weighted predictions.}
\label{fig:architecture}
\end{figure*}
\IEEEPARstart{H}{and} pose estimation is a fundamental and challenging problem in computer vision,
and it has a wide range of vision applications such as human-computer interface
(HCI)~\cite{sato2001real,erol2007vision} and augmentation reality (AR)~\cite{barsoum2016articulated}. So far, there
have been a number of methods~\cite{Ge_2016_CVPR,deng2017hand3d,Ge_2017_CVPR,
sinha2016deephand,sun2015cascaded, wan2016direction,ye2016spatial,li20153d,zhang2016learning,
wang2009real,romero2009monocular,Zimmermann_2017_ICCV,panteleris2017back,oberweger2017deepprior++}
proposed on this topic, and significant progress has been made with the emergence of deep
learning and low-cost depth sensors. The performance of previous RGB image based methods
is limited to the clustered background, while the depth sensor is able to generate depth images in low
illumination conditions, and it is simple to do hand detection and hand segmentation.
Nevertheless, it is still difficult to accurately estimate 3D hand pose in the practical scenarios due
to data noises, high freedom of degree for hand motion, and self-occlusions between fingers.

The typical pipeline for estimating hand joints from depth images could be separated into two stages: 1) extracting robust features
from depth images; 2) regressing hand pose based on the extracted features. Current approaches concentrate on improving the different
algorithms in each of the stages. Specifically, in~\cite{Ge_2016_CVPR}, the features
extracted by multi-view CNNs are utilized for regression, then the same author in~\cite{Ge_2017_CVPR} replaces the 2D CNN by 3D CNN
to fully exploit 3D spatial information. Besides, hierarchical hand joint regression~\cite{ye2016spatial} and iterative
refinement~\cite{oberweger2015hands,oberweger2015training} for hand pose are deployed in several approaches.

In this paper, we propose a novel method named as \textbf{CADSTN} (\textbf{C}ontext-\textbf{A}ware \textbf{D}eep \textbf{S}patio-\textbf{T}emporal \textbf{N}etwork)
to jointly model the spatio-temporal context for 3D hand pose estimation. We adopt sliced 3D volumetric representation, which keeps the structure of the hand to explicitly model
the depth-aware spatial context. Moreover, motion coherence between the successive frames is
exploited to enforce the smoothness of the predictions of the images in sequence. The model is able to learn the representations of the spatial
information and the temporal structure in the image sequences. Moreover, the model is capable of combining the spatio-temporal
properties for final prediction. As shown in Figure~\ref{fig:architecture}, the architecture is
composed of three parts: first, we sufficiently capture the spatial information via \emph{Spatial Network}; second, the dependency between
frames is modeled via a set of LSTM nodes in \emph{Temporal Network}; third, to exploit spatial and temporal context simultaneously,
the predictions from the above two networks are utilized for the final prediction via \emph{Fusion Network}. The main contributions of our work are summarized as the following two folds:
\begin{itemize}
  \item
  First, we propose a unified spatio-temporal context modeling approach with group input and group output, which benefits from the inter-image
  correspondence structures between consecutive frames. The proposed approach extracts the feature representation for both unary image and
  successive frames, which generally leads to the performance improvement of hand pose estimation.
  \item
  Second, we present an end-to-end neural network model to jointly model the temporal dependency relationships among the multiple images while
  preserving spatial information for an individual image in a totally data-driven manner. Moreover, an adaptive fusion method is enabled to make
  the network dynamically adapt to the various situations.
\end{itemize}

We evaluate our proposed method on two public datasets: NYU~\cite{tompson2014real} and ICVL~\cite{sun2015cascaded}.
With detailed analysis of the effects of different components of our method, and the
experimental results demonstrate that most results of our method are best
comparing with state-of-the-art methods. Our method runs in 60fps on a single
GPU that satisfies the practical requirements\footnote{A demo video is available online at
\url{https://goo.gl/tbnWse}}.

The rest of the paper is organized as follows. In Section~\ref{sec:related work}, we review some works for 3D hand pose estimation.
In Section~\ref{sec:hand pose estimation}, we introduce our proposed method. In Section~\ref{sec:experiments},
we present the experimental results of the comparison with state-of-the-art methods and ablation study.
Finally, the paper is concluded in Section~\ref{sec:conclusion}.

\section{Related Work}\label{sec:related work}
\subsection{Feature Representation Learning}\label{sec:representation learning}
Most state-of-the-art performance hand pose estimation methods based on deep learning rely on visual
representation learned from data. Given a depth image, different methods try to learn the
sufficient context and get the robust feature. For example, in~\cite{oberweger2015hands}, the original segmented
hand image is downscaled by several factors, and each scaled image is followed with a neural
network to extract multi-scale feature. Although increasing the number of scale factor may
improve the performance, the computational complexity will also be increased.
In~\mbox{\cite{guo2017region}}, Guo et al. propose a regional feature ensemble neural network and extend it into
3D human pose estimation in~\mbox{\cite{wang2018region}}, Chen et al.~\mbox{\cite{chen2017pose}} propose a cascaded
framework denoted as Pose-REN to improve the performance of region ensemble network. In~\mbox{\cite{Moon_2018_CVPR_V2V-PoseNet}},
various combinations of voxel and pixel representations are experimented, and finally use a voxel-to-voxel prediction framework
to estimate per-voxel likelihood.

Furthermore, recovering the fully 3D information catches researcher's attention. Multi-View
CNN~\cite{Ge_2016_CVPR} is introduced for hand pose estimation, which exploits depth cues to recover 3D
information of hand joints. In~\cite{deng2017hand3d}, 3D CNN and truncated signed
distance function (TSDF)~\cite{newcombe2011kinectfusion} are adopted to learn both the
3D feature and the 3D global context. Ge et al.~\cite{Ge_2017_CVPR} take advantage of the above
two methods, and propose a multi-view 3D CNN to regress the hand joint locations.
In addition, data augmentation is usually considered for extracting robust feature. A synthesizer
CNN is used to generate synthesized images to predict an estimate of the 3D pose by using a feedback
loop in~\cite{oberweger2015training}. Wan et al.~\cite{Wan_2017_CVPR} use two generative models,
variational autoencoder (VAE)~\cite{kingma2013auto} and generative adversarial network
(GAN)~\cite{goodfellow2014generative}, to learn the manifold of the hand pose, with an alignment
operation, pose space is projected into a corresponding depth map space via a shared latent
space. Our method is different from the aforementioned architectures in the following ways: 1) we
deploy the sliced 3D volumetric representation to recover 3D spatial information. 2) we extract the
temporal property from an image sequence to enhance the consistency between images.

\subsection{Spatio-Temporal Modeling}\label{sec:spatio-temporal modeling}
Towards the tasks based on the video, spatio-temporal modeling
is widely used (e.g. hand tracking, human pose tracking and action recognition).

In the matters of hand tracking problems, Franziska et al.~\mbox{\cite{mueller2017real}} propose a kinematic pose tracking energy to estimate the joint angles and to overcome the challenge occlusion cases. In~\mbox{\cite{oberweger2016efficiently}}, Oberweger et al. propose a semi-automated method to annotate the hand pose in an image sequence by exploiting spatial, temporal and appearance constraints.

In the context of human pose tracking,
Fragkiadaki et al.~\cite{fragkiadaki2015recurrent} propose an Encoder-Recurrent-Decoder (ERD)
module to predict heat maps of frames in an image sequence. More recently, Song et al.~\cite{Song_2017_CVPR}
introduce a spatio-temporal inference layer to perform message passing on general loopy spatio-temporal graphs.

In terms of action recognition problem, Molchanov et al.~\cite{molchanov2016online} employ RNN and deep 3D CNN on a whole
video sequence for spatio-temporal feature extraction. Our approach is closely related to~\cite{simonyan2014two}
who propose a two-stream architecture to encode the spatial and temporal information separately. Our approach
is based on the depth image, which leads us to learn spatio-temporal feature via LSTM instead of optical flow.

\subsection{Regression Strategies}\label{sec:regression}
Regression strategy has the close relationship with the performance of methods, and it can be roughly classified
into three camps: 1) 2D heat-map regression and 3D recovery; 2) directly 3D
hand joint coordinates regression; 3) cascaded or hierarchical regression. In~\cite{tompson2014real, Ge_2016_CVPR},
2D heat-maps, which represent the 2D location for hand joints, are regressed based on the extracted feature,
then the 3D positions are determined via recovery.
Directly 3D hand joint coordinates regression predicts the hand locations in a single forward-pass,
which exhibits superior performances comparing to the previous 2D heat-map regression~\cite{
zhou2016model,Ge_2017_CVPR,sinha2016deephand,deng2017hand3d,Wan_2017_CVPR,guo2017region}.

Cascaded and hierarchical regression has shown good performance in hand pose estimation. Oberweger
et al.~\cite{oberweger2015hands} use the iterative refinement stage to increase the accuracy, and
in~\cite{oberweger2015training}, they use the generated synthesized images to correct the initial
estimation by using a feedback loop. Tang et al.~\cite{tang2014latent,tang2015opening} introduce
hierarchical regression strategies to predict tree-like topology of hand. Similarly, the strategy to
first estimate hand palm and sequentially estimate the remaining joints is adopted
in~\cite{sun2015cascaded,wan2016direction}. Recently, Ye et al.~\cite{ye2016spatial} combine the
attention model with both cascaded and hierarchical estimation, and propose an end-to-end learning
framework.
\section{Hand Pose Estimation}\label{sec:hand pose estimation}
Figure~\ref{fig:architecture} illustrates the main building blocks of our method.
Given an image sequence, two predictions are estimated by \emph{Temporal Network} and \emph{Spatial Network},
then the predictions are fused by the output from \emph{Fusion Network}.

\subsection{Overview}\label{sec:overview}
We denote a single depth image as $\I$, and the corresponding $K$ hand joints are represented as a vector
$\J=\{\j_k\}_{k=1}^K \in \mathbb{R}^{3K}$, where $\j_k$
is the 3D position of $k$-th joint in the depth image $\I$.

Currently, most discriminative approaches concentrate on modeling a mapping function $F$ from the input image to hand pose.
We define a cost function as follows:
\begin{equation}\label{eq:cost}
C(F, \J) = \frac{1}{N}\sum_{i=1}^N\|F(I_i) - \J_i\|_2
\end{equation}
where $\|\cdot\|_2$ calculates the $l_2$ norm, $N$ is the number of images. Then the expected mapping function $\hat{F}$ is \
expressed as $\hat{F} = argmin_{F \in \mathcal{F}} C(F, \J)$, where $\mathcal{F}$ is the hypothesis space.

To obtain a robust mapping function, most approaches concentrate on the hand structure to capture features for estimation.
In this paper, we capture the spatial and temporal information simultaneously to learn features for prediction.
As shown in Figure~\ref{fig:architecture}, the architecture is separated into three parts.
The first part is denoted as \emph{Spatial Network}, which is applied on a single frame to learn a mapping function $F_{spa}$.
Inspired by the texture based volume rendering used in the medical image~\cite{hopf1999accelerating}, we slice the depth
image into several layers, which makes the different joints scatter in the different layer. We feed the depth image and
corresponding sliced 3D volumetric representation into the network, depth and spatial information are extracted hierarchically
for hand pose estimation.

The second part is denoted as \emph{Temporal Network}. This network concentrates on the time coherence property in image sequences,
and learns the mapping $F_{temp}$ from an image sequence to a set of hand poses. By using LSTM layer,
this network takes the features from previous frames into account when estimating the pose in the current frame.

The third part is named as \emph{Fusion Network}. For the sake of predicting pose via the combination of information from the
individual and consecutive frames, we fuse the aforementioned two networks as an integration by adaptive fusion method,
the final prediction is calculated as the summation
of the different \textit{network}s' outputs. By means of employing this approach,
spatial information from the individual frame and temporal
property between frames are thought to have in mind as a unified scheme.

\subsection{Proposed Method}\label{sec:proposed method}
In this section, we present details of our method for hand pose estimation.\vspace{0.5em}

\noindent \textbf{Spatial Network}\vspace{0.5em}\label{sec:spatial network}\\
\indent In an individual frame, the spatial information is critical for hand pose estimation.
By adopting 3D sliced volumetric representation and Deeply-Fusion network~\cite{Chen_2017_CVPR}, we extract features
to learn a robust mapping function.

\begin{figure}[htbp]
\centering
    \includegraphics[width=1\linewidth]{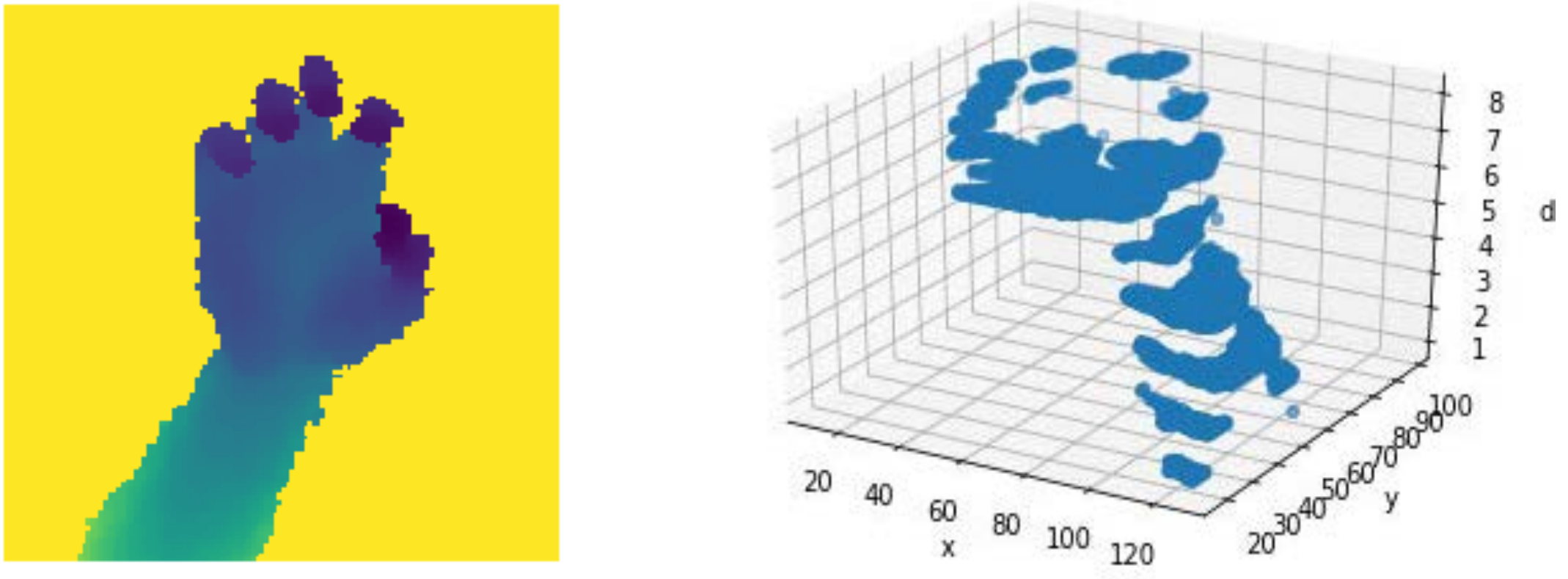}
    \caption{Depth image and sliced 3D volumetric representation.}
\label{fig:sliced 3D volumetric representation}
\end{figure}

In the original depth image $\mathit{I}$, the value of the pixel encodes the depth information.
We first crop the hand region out of the image, this step is same as~\cite{oberweger2015hands}.
As shown in Figure~\ref{fig:sliced 3D volumetric representation}, the cropped depth image is
resized to $M \times M$, and depth $D(x,y)$ is represented as a function of $(x,y)$ coordinates.
Previous methods represent the 3D hand in different ways.
In~\cite{III_2015_ICCV}, the depth image is recovered as a binary voxel grid for training. The hand surface and
occluded space~\cite{wu20153d} are filled with $1$, the free space is filled with $0$.
In~\cite{deng2017hand3d,Ge_2017_CVPR}, TSDF and projective Directional TSDF (D-TSDF) are adopted to
encode more information in the 3D volumetric representation. However, the size of 3D volume used in
these methods is commonly set as $32 \times 32$.
Different from previous methods, we employ a solution
that the hand surface is sliced into $L$ pieces. Then our sliced 3D volumetric
representation is an $M \times M \times L$ voxel grid, where $L$ represents the number of sliced layers
in $z$-axes. For an exact cropped depth image, the free space and occluded space are omitted, then we denote
$D_{min}$  and $D_{max}$ as the depth of closest and farthest surface point, then we divide $[D_{min}, D_{max}]$
equally into $L$ pieces. So the binary sliced 3D volumetric representation $V[x,y,l]$ is filled with $1$ if
the depth value $D(x,y)$ is in the range of $[D_{min} + \frac{D_{max} - D_{min}}{L}l, D_{min} +
\frac{D_{max} - D_{min}}{L}(l+1)]$, where $l=0, 1,\dots ,L-1$. In Figure~\ref{fig:sliced 3D volumetric representation},
from the top layer to the bottom layer, the fingertips and palm center scatter in the different layers.

\begin{figure}[htbp]
    \centering
    \includegraphics[width=1\linewidth]{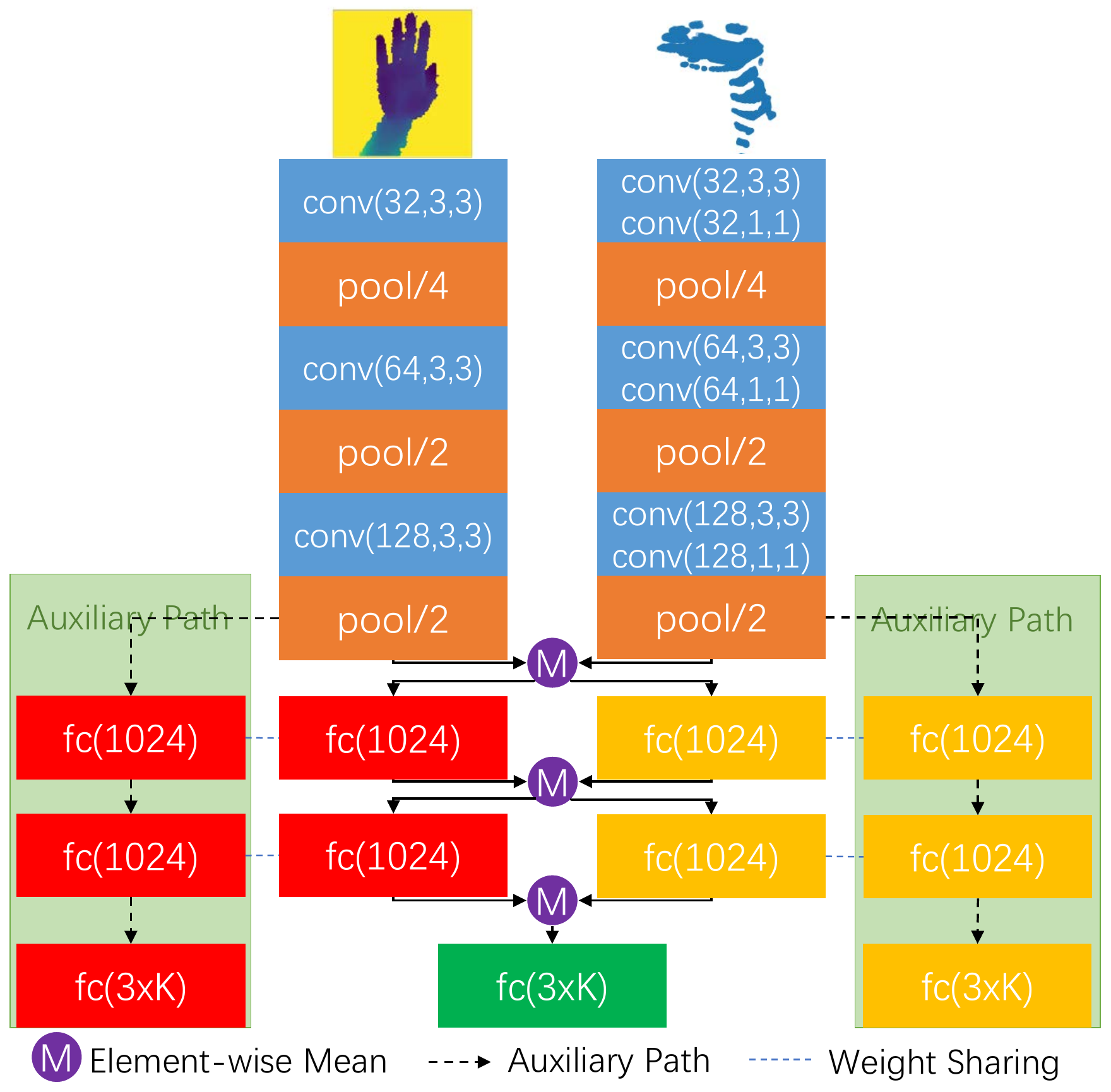}
    \caption{\emph{Spatial Network}. Depth image and sliced 3D volumetric representation are input
    into \emph{Spatial Network}, the different features are fused together hierarchically. For the
    purpose of obtaining more representative capability for each input, the auxiliary paths are added
    in training stage, and the layers in the auxiliary paths are same as the main network. In the training
    stage, the total losses are consist of three losses, i.e. three regression loss in auxiliary paths
    and main network. When testing, the auxiliary pathes are removed. For convolution layer, the number
    in the parenthesis is the kernel size. For pooling layer, the number is stride. For fully connected
    layer, the number is the length of output feature.}
\label{fig:spatial network}
\end{figure}

Sliced 3D volumetric representation keeps the structure of the hand.
To obtain the hand structure and depth information simultaneously, we input depth image and sliced 3D volumetric
representation at the same time. As shown in Figure~\ref{fig:spatial network}, the left part extracts features from
depth image, and the right part extracts features from sliced 3D volumetric representation. We denote the features
from max-pooling layers as $\phi_{depth}$ and $\phi_{3D}$, then the features are hierarchically fused by Deeply-Fusion
Network~\cite{Chen_2017_CVPR} as follows:
\begin{equation}\label{eq:deep fusion}
\begin{aligned}
&\phi_0 = \phi_{depth} \oplus \phi_{3D} \\
&\phi_m = \mathit{H}_m^{depth}(\phi_{m-1}) \oplus \mathit{H}_m^{3D}(\phi_{m-1}) \\
&m=1, 2.
\vspace{1em}
\end{aligned}
\end{equation}
where $\oplus$ is element-wise mean for features, $\mathit{H}$ is the transformation for features learned by a
fully connected layer, $l$ is the index for layer.

Owing to the adoption of several fully connected layers, the network tends to overfitting, so we use
an auxiliary loss as the regularization as well as dropout. In Figure~\ref{fig:spatial network},
every fully connected layer is followed by a dropout layer, the nodes are dropout randomly
with 30\% probability. For the purpose of obtaining more representative capability for each input,
the auxiliary paths are added in training stage. As shown in the green boxes, the layers in the
auxiliary paths are same as the main network, the layers connected by the blue dotted line
share parameters. And in the training stage, the total losses are consist of three losses,
i.e. three regression losses in auxiliary paths and main network. When testing, the auxiliary
paths are removed.\vspace{0.5em}

\vspace{1em}
\noindent \textbf{Temporal Network}\vspace{0.5em}\label{sec:temporal netowork}

As we know, the hand poses between successive frames are closely related (e.g. when grabbing, the joints
on fingers are most likely to be closer). The temporal property in the between frames are also critical for
estimating hand joints. So we extend the mapping function from single frame to a sequence of images. The
sequential prediction problem could be reformulated as follows:
\begin{equation}\label{eq:seq2seq}
(\J_1, \dots, \J_{T}) = F_{temp}(\I_1, \dots, \I_{T})
\end{equation}
where $T$ is the number of images in a sequence.

\begin{figure}[htbp]
\centering
    \includegraphics[width=0.5\linewidth]{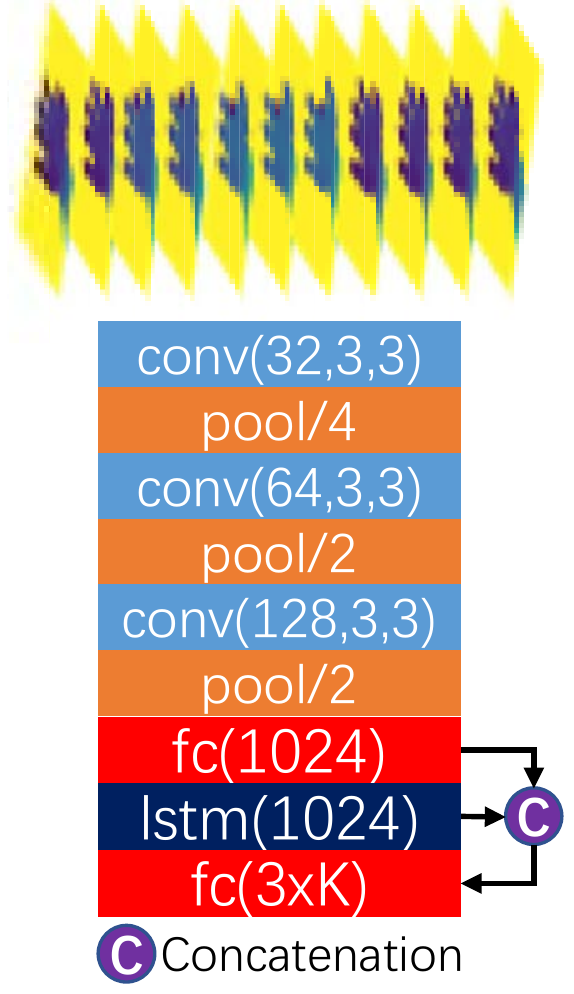}
    \caption{\emph{Temporal Network}. It learns a mapping function $F_{temp}$ to process an image sequence,
    a set of images input and LSTM layer captures the temporal context in these images. For LSTM layer, the number
    in the parenthesis is the dimension of the hidden state feature.}
\label{fig:temporal network}
\end{figure}

Currently, due to the feedback loop structure in the recurrent neural network (RNN), it possesses powerful "long-term dependencies"
modeling capability, and is widely used in computer vision tasks. In general, RNN is difficult to train due to the vanishing
gradient and error blow-up problems~\cite{kawakami2008supervised}. In this paper, we adapt LSTM~\cite{hochreiter1997long,zaremba2014learning},
a variant of RNN which is expressive and easy to train, to model the temporal context of successive frames.
As shown in Figure~\ref{fig:temporal network}, an image sequence is taken as input and the network gives out the estimated poses.
Without loss of generality, we think about the $t$-th depth image $I_t$,
the image is fed into the convolution neural network and the feature $\phi_t$ extracted from the first fully connected layer is denoted as follows:
\begin{equation}
\phi_t=\H_{\theta_c}(I_t).
\end{equation}

Then we feed the features $\phi_t$ into LSTM layer, and get the hidden state $h_t$ as the new features.
\begin{equation}
\begin{aligned}
&i_t := \sigma(W_{hi} * h_{t-1} + W_{xi} * \phi_t + b_i) \\
&f_t := \sigma(W_{hf} * h_{t-1} + W_{xf} * \phi_t + b_f) \\
&o_t := \sigma(W_{ho} * h_{t-1} + W_{xo} * \phi_t + b_o) \\
&c_t := (f_t \odot c_{t-1}) + (i_t \odot tanh(W_{hc} * h_{t-1} + W_{xc} * \phi_t + b_c)) \\
&h_t := o_t \odot tanh(c_t)
\end{aligned}
\end{equation}
where $\sigma(\cdot)$ is sigmoid function, $tanh(\cdot)$ is tanh function, $\odot$ is element-wise product, and matrices
$W_{h*}, W_{x*}, b_{*}$ ($*$ is in $i, f, o, c$) are parameters for the gates in LSTM layer.
The features $h_t$ output from LSTM layer and the original features $\phi_t$ are concatenated,
and then are fed into the last fully connected layer to regress the hand pose coordinates.\vspace{0.5em}

\begin{figure}[t]\footnotesize
\centering
    \includegraphics[width=0.75\linewidth]{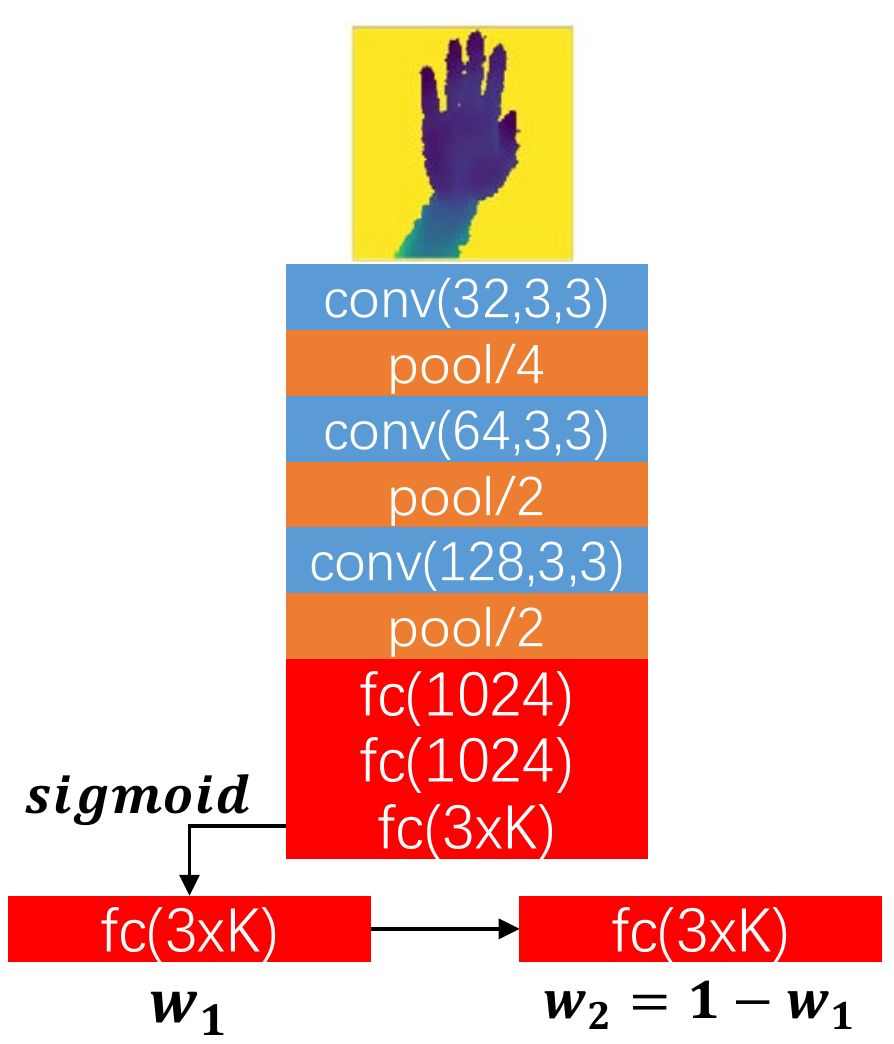}
    \caption{\emph{Fusion Network}, the network outputs weights for two predictions from \emph{Spatial Network}
    and \emph{Temporal Network}. $\bm{w}_1$ and $\bm{w}_2$ are learned as the confidence of two predictions.}
\label{fig:fusion}
\end{figure}

\noindent \textbf{Fusion Network}\vspace{0.5em}\label{sec:fusion network}\\
\indent The aforementioned \emph{Temporal Network} and \emph{Spatial Network} estimate the joints by placing emphasis on
capturing spatial and temporal information, we name the predictions from these two networks as $\J_{temp}$ and $\J_{spa}$ respectively.
Due to the importance of spatial and temporal information, we jointly model the spatio-temporal properties,
which adaptively integrates different predictions for the final estimation. As shown in Figure~\ref{fig:fusion},
\emph{Fusion Network} uses an activation function $sigmoid$ after the last fully connected layer, then output $\bm{w}_1$ and $\bm{w}_2$.
\emph{Fusion Network} fuses two predictors, and gives out the final prediction as follows:
\begin{equation}\label{eq:Fusion Network}
\centering
\J_{out} = \bm{w}_1 \odot \J_{temp} + \bm{w}_2 \odot \J_{spa}
\end{equation}
where $\bm{w}_1 + \bm{w}_2 = \bm{1}$, and $\bm{1} \in \mathbb{R}^{3K}$ is a vector that all elements are one.
The weights $\bm{w}_1$ and $\bm{w}_2$ are learned as the confidence of two predictions, and the final prediction $\J_{out}$
is estimated as the weighted summation of each prediction.

Because \emph{Temporal Network} considers the temporal information and \emph{Spatial Network} extracts
the spatial information, then the network infers the hand joint locations depend on spatial and temporal features.

Finally, we summarize our proposed method in Algorithm~\ref{alg:algorithm}.

\begin{algorithm}[htbp]\footnotesize
\caption{Summary for Context-Aware Deep Spa-tio-Temporal Network}
\label{alg:algorithm}
    \KwIn{Training data $D = \{\I_i, \J_i\}_{i=1}^N$, testing images $D' = \{\I_j\}_{j=1}^{N'}$, number of LSTM cells in LSTM layer $T$}
    \KwOut{The predicted hand poses $\{\J_j\}_{j=1}^{N'}$}
\tcc{Training stage}
    \emph{Spatial Network} is trained with $D$, and learn the mapping function $F_{spa}$ for an individual frame to minimize Equation~\ref{eq:cost}\;
    Train \emph{Temporal Network} to learn the mapping function $F_{temp}$ in Equation~\ref{eq:seq2seq} for a set of images $F_{temp}:\{\I_1, \dots, \I_T\} \rightarrow \{\J_1, \dots, \J_T\}$\;
    Fix the parameters in the above two networks, then train the parameters in \emph{Fusion Network} to learn parameters to output confidential weights $\bm{w}_1$ and $\bm{w}_2$ in Equation~\ref{eq:Fusion Network}\;
\tcc{Testing stage}
    Forward $T$ testing images into trained network, get the predictions $\J_{spa}$ and $\J_{temp}$ from \emph{Spatial Network} and \emph{Temporal Network}\;
    Calculate the prediction via Equation~\ref{eq:Fusion Network}.
\end{algorithm}

\section{Experiments}\label{sec:experiments}
In this section, we evaluate our method on two public datasets for comparison with the state-of-the-art methods.
In addition, we do ablation experiments and analyse the performance of each component.

\begin{figure*}[t]\footnotesize
\centering
    \subfloat[]{
        \label{fig:comparison NYU}
        \begin{minipage}[t]{240pt}
            \centering
            \includegraphics[width=1.\linewidth]{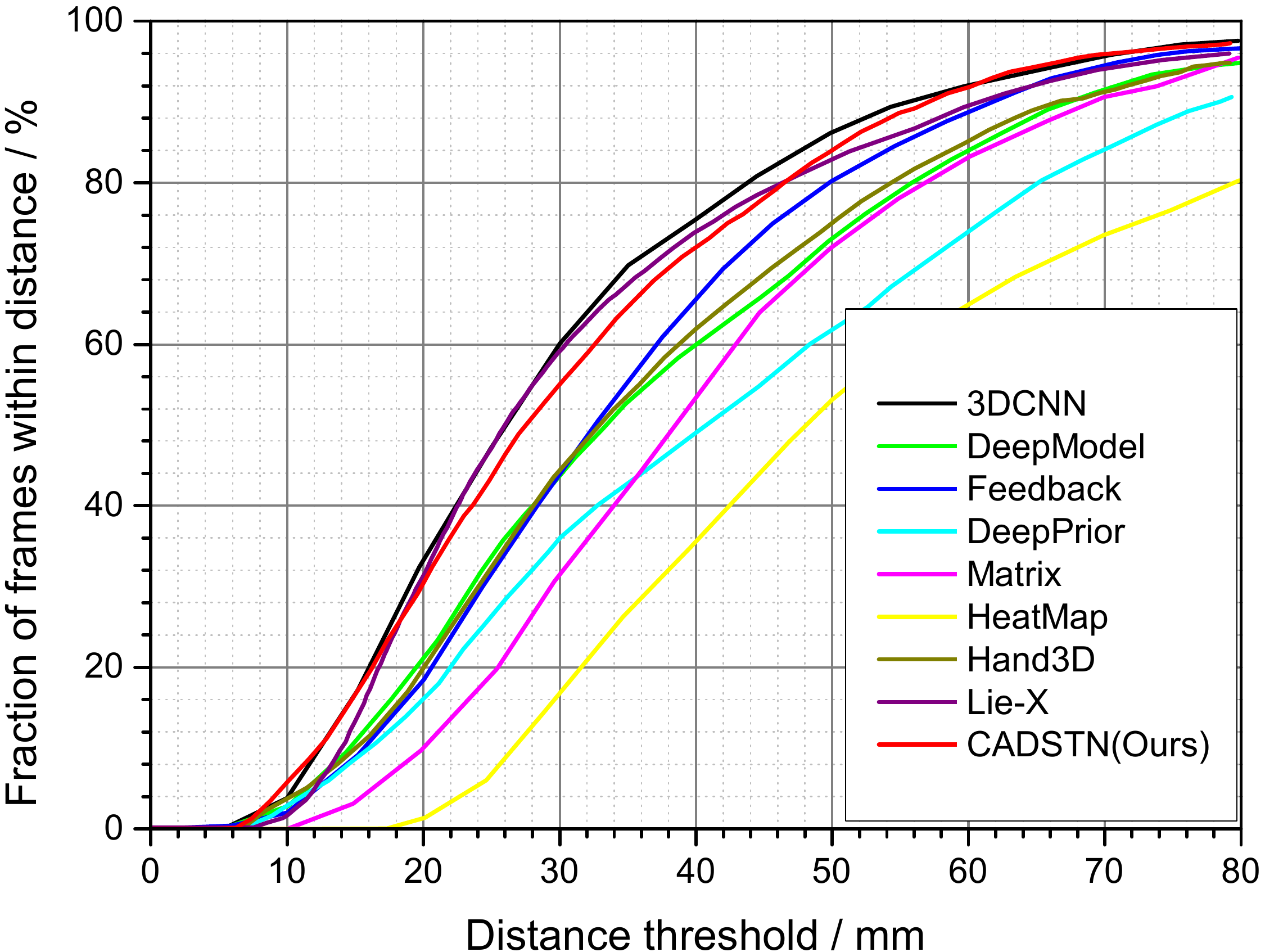}
        \end{minipage}
    }
    \subfloat[]{
        \label{fig:comparison ICVL}
        \begin{minipage}[t]{240pt}
            \centering
            \includegraphics[width=1.\linewidth]{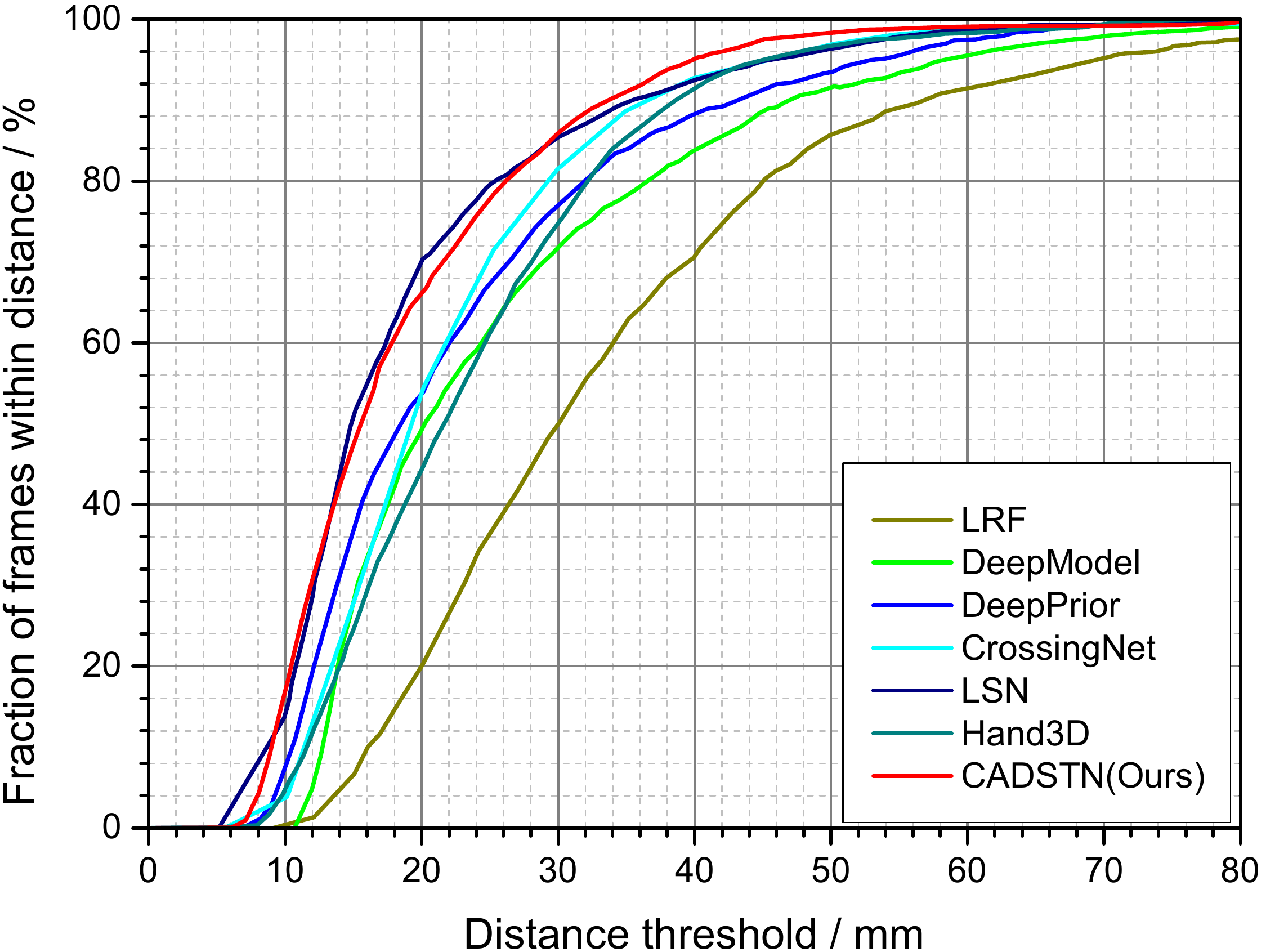}
        \end{minipage}
    }
    \\
    \subfloat[]{
        \label{fig:joint mean NYU}
        \begin{minipage}[t]{244pt}
            \centering
            \includegraphics[width=1\linewidth]{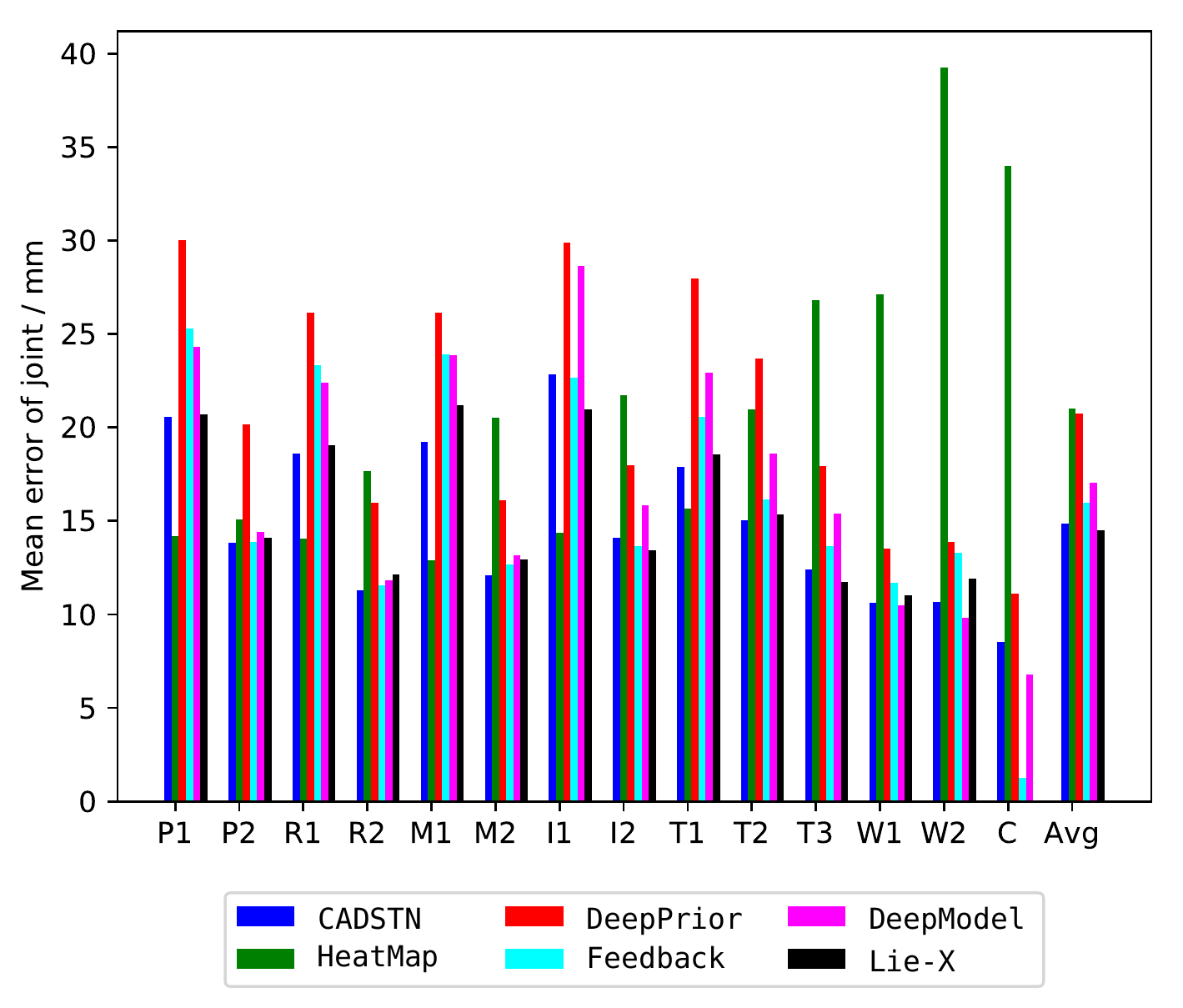}
        \end{minipage}
    }
    \subfloat[]{
        \label{fig:joint mean ICVL}
        \begin{minipage}[t]{250pt}
            \centering
            \includegraphics[width=1\linewidth]{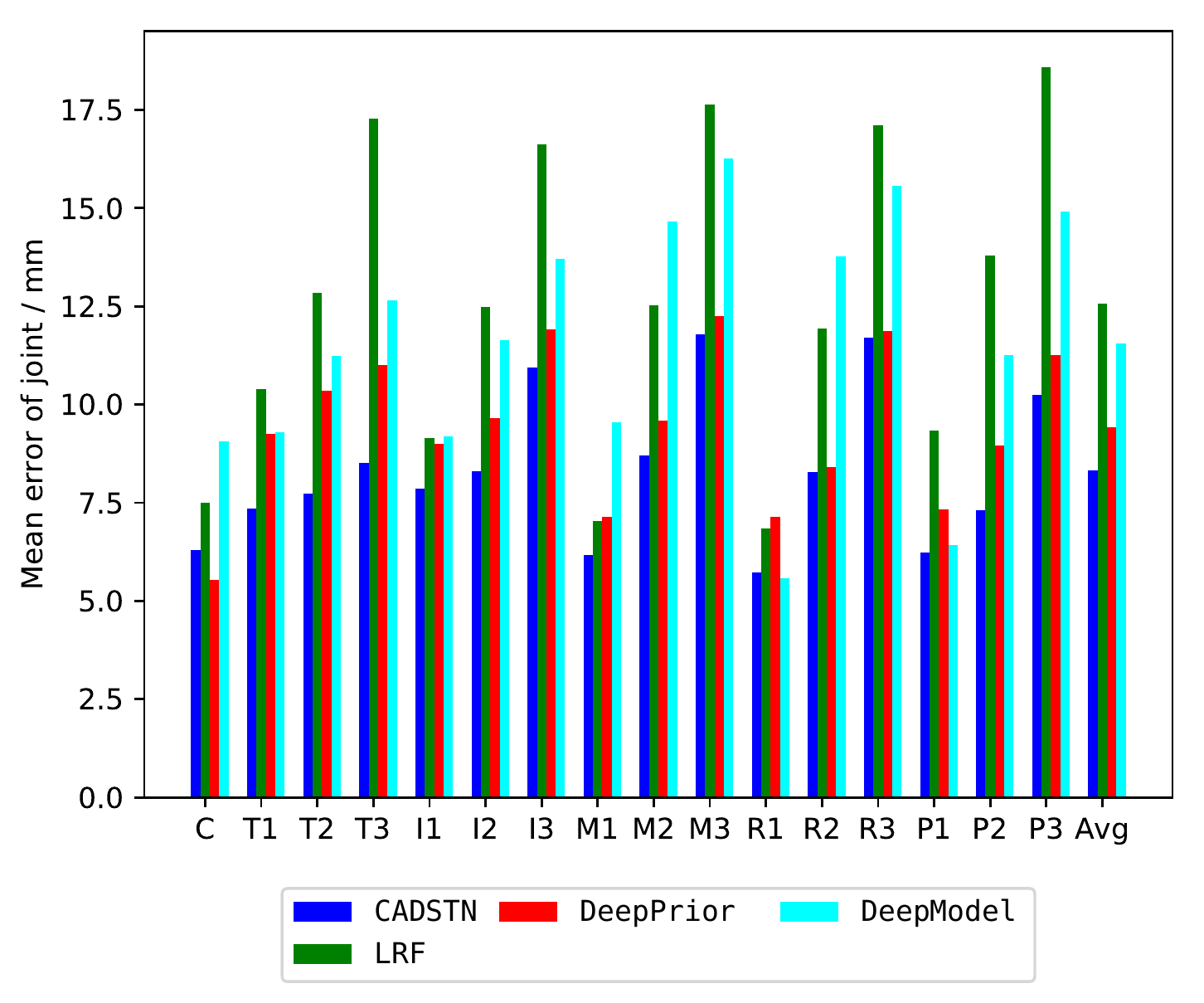}
        \end{minipage}
    }
    \caption{Comparison with state-of-the-art methods on NYU dataset and ICVL dataset. Figure(a) and Figure(b): the
    proportion of good frames over different thresholds on NYU dataset and ICVL dataset. Figure(c) and Figure(d):
    per-joint error distance on NYU dataset and ICVL dataset. The palm and fingers are indexed as C: palm, T:
    thumb, I: index, M: middle, R: ring, P: pinky, W: wrist.}
    \label{fig:Results}
\end{figure*}

\subsection{Experiment Setting}\label{sec:experiment setting}
\subsubsection{Datasets}\label{sec:datasets}
We evaluate our method on NYU~\cite{tompson2014real} and ICVL~\cite{tang2014latent}, details about the datasets
are summarized in Table~\ref{tab:datasets}, where \textbf{Train} is the number of train images, \textbf{Test}
is the number of test images and the number in bracket is the number of sequences; \textbf{Resolution} is the resolution
of images; \textbf{Annotation} is the number of annotated joints.
NYU dataset is challenging because of its wider pose variation and noisy image as well as limited annotation accuracy.
On ICVL dataset, the scale of image is small and the discrepancies between training and testing is large, these make
the estimation difficult.

\begin{table}[htbp]
\centering
\caption{Dataset for experiments.}
\begin{tabular}{c|c|c|c|c}
    \hline
    \textbf{Dataset} & \textbf{Train}     & \textbf{Test}    & \textbf{Resolution} &\textbf{Annotation}\\
    \hline
    NYU     & 72k (1)   & 8k(2)   & $640\times480$  & 36\\
    ICVL    & 22k (10)  & 1.6k(2) & $320\times240$  & 16\\
    \hline
\end{tabular}
\label{tab:datasets}
\end{table}

\subsubsection{Evaluation Metrics}\label{sec:evaluation metrics}
The evaluation follows the standard metrics proposed in~\cite{oberweger2015hands}, including accuracy,
pre-joint error distances and average error distance. As mentioned above, we denote $\j_{kn}$ as
the predicted $k$-th joint location in the $n$-th frame, $\j_{kn}^{*}$ is the corresponding ground
truth. The total number of frames is denoted as $N$, and $K$ is the number of hand joints.

Per-joint error distance calculates the average Euclidean distance between the predicted
joint location and the ground truth in 3D space.
\begin{equation}
err_k = \frac{\sum_{n=1}^N(\|\j_{kn} - \j_{kn}^{*} \|)}{N}
\end{equation}

Average error distance computes the mean distance for all joints.
\begin{equation}
ave = \frac{\sum_{k=1}^K{err_k}}{K}
\end{equation}

Accuracy is the fraction of frames that all predicted joints from the ground truth in a
frame below the given distance threshold $\delta$.
\begin{equation}
acc_{\delta} = \frac{\mathbbm{1}(max_k(\| \j_{kn} - \j_{kn}^{*}\|) \leq \delta)}{N}
\end{equation}
where $\mathbbm{1}(\cdot)$ is the indicator function.

\subsubsection{Implementation Details}\label{sec:implementation}
We implement the training and testing with Caffe\cite{jia2014caffe} framework. The pre-process step
follows~\cite{oberweger2015hands}, then the cropped image is resized to $128 \times 128$ with the
depth value normalized to [-1, 1], and for sliced 3D volumetric representation, $L$ is set to 8
in the experiments. Moreover, we augment the dataset from the available dataset.

\textbf{Data augmentation}
On NYU dataset, we do augmentation by random rotation and flipping. On one side, we randomly rotate the images
by an angle selected from $90^\circ$ $-90^{\circ}$ and $180^{\circ}$. On the other side, we flip the image
vertically or horizontally. For each image in the dataset, we just generate one image by the above-mentioned augmentation
method. So the size of the dataset is twice as many as the original. On ICVL dataset, in-plane rotation has been applied and
there are 330k training samples in total, so the augmentation is not used on ICVL.

\textbf{Training configuration}
Theoretically, training the three networks jointly is considerable, but it takes a longer training time to
optimize the parameters of the three networks. Besides, we have tried to fine-tune the spatial and temporal network,
but there is no much difference to freeze the Spatial Network and Temporal Network due to the limited size of
data set. Practically, we strike a balance between training complexity and efficiency, so we adopt the simpler method as follows. We train
\emph{Temporal Network} and \emph{Spatial Network} first, then train \emph{Fusion Network} by fixing the above two networks.
The training strategy is same for \emph{Spatial Network} and \emph{Temporal Network}, we optimize the parameters by using back-propagation
and apply Adam~\cite{kingma2015adam} algorithm, the batch size is set as 128, the learning rate starts with 1e-3 and decays every 20k iterations.
\emph{Spatial Network} and \emph{Temporal Network} are trained from scratch with 60k iterations.
\emph{Fusion Network} is trained by 40k iterations with the above two networks fixed.
Our training takes place on machines equipped with a 12GB Titan X GPU.

\subsection{Experimental Results}\label{sec:experimental results}
\subsubsection{Comparison with State-of-the-art}\label{sec:comparison}
We compare our method with several state-of-the-art hand pose estimation methods~\cite{tompson2014real,Wan_2017_CVPR,
oberweger2015hands,oberweger2015training,zhou2016model,tang2014latent,Ge_2017_CVPR,sinha2016deephand}. What worth mentioning is
that some methods provide predicted joints but some not, we calculate the metrics of some methods~\cite{tompson2014real,
oberweger2015hands,oberweger2015training,zhou2016model,tang2014latent} available online and estimate the others from the
figures in their papers.

\begin{figure}[t]
  \centering
  \includegraphics[width=0.4\linewidth]{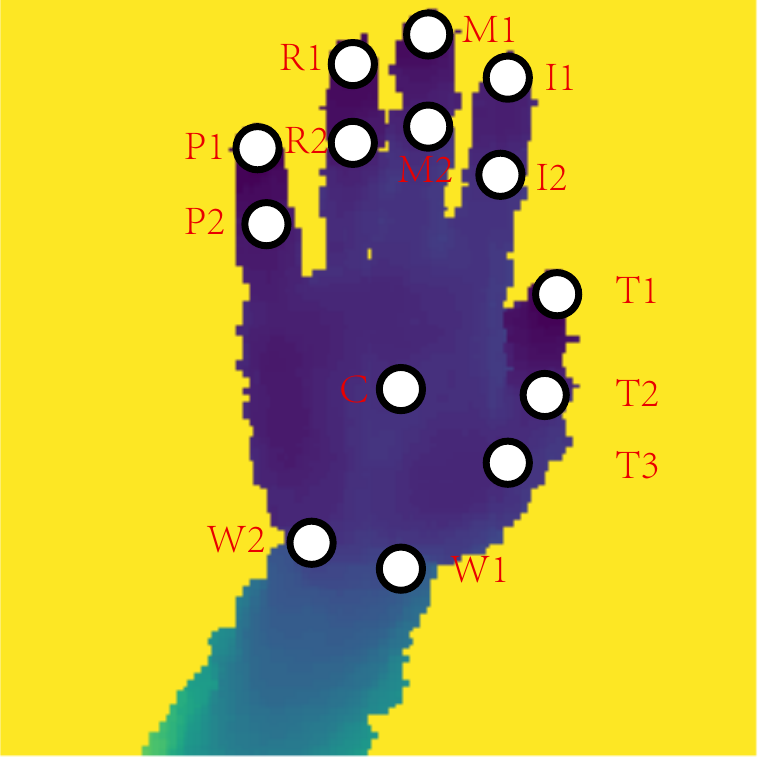}\\
  \caption{Hand annotation used on NYU dataset.}\label{fig:hand NYU}
\end{figure}

\textbf{NYU Dataset} contains 72k training images from one subject and 8k testing images from two subjects. There are totally 36
annotated joints, and we only evaluate a subset of 14 joints as used in other papers for a fair comparison. The annotation for hand is illustrated in Figure~\ref{fig:hand NYU}.

We compare our method with several state-of-the-art approaches on NYU dataset, including 3DCNN~\cite{Ge_2017_CVPR}, DeepModel~\cite{zhou2016model},
Feedback~\cite{oberweger2015training}, DeepPrior~\cite{oberweger2015hands}, Matrix~\cite{sinha2016deephand}, HeatMap~\cite{
tompson2014real}, Lie-X~\cite{xu2017lie}.

The accuracy and the pre-joint error distances are shown in Figure~\ref{fig:Results},
our proposed method has a comparable performance with state-of-the-art methods.
In Figure~\ref{fig:comparison NYU}, our method outperforms the majority of methods. For example, the proportion of good frames is about 10\%
higher than Feedback when distance threshold is 30mm. 3DCNN adopts the augmentation and projective D-TSDF method, and trains
a 3D convolution neural network which maps the sliced 3D volumetric representation to 3D hand pose, our method could not as accurate as 3D
CNN from 15-60mm due to the sufficient 3D spatial information captured by the 3D CNN.
Lie-X infers the optimal hand pose in the manifold via a Lie group based method, it stands out over the range of 20-46mm, but our method
overtakes at other threshold.
In Figure~\ref{fig:joint mean NYU}, the per-joint error distance for five methods and our method are illustrated.

Table~\ref{tab:mean error NYU} reports the average error distance for different methods. Generally, results show that our method has a
comparable performance with competing methods. Our method outperforms~\cite{tompson2014real,oberweger2015hands,oberweger2015training,zhou2016model}
by a large margin, and is comparing to~\cite{xu2017lie}.

\begin{figure*}[htbp]\footnotesize
\centering
    \includegraphics[width=1\linewidth]{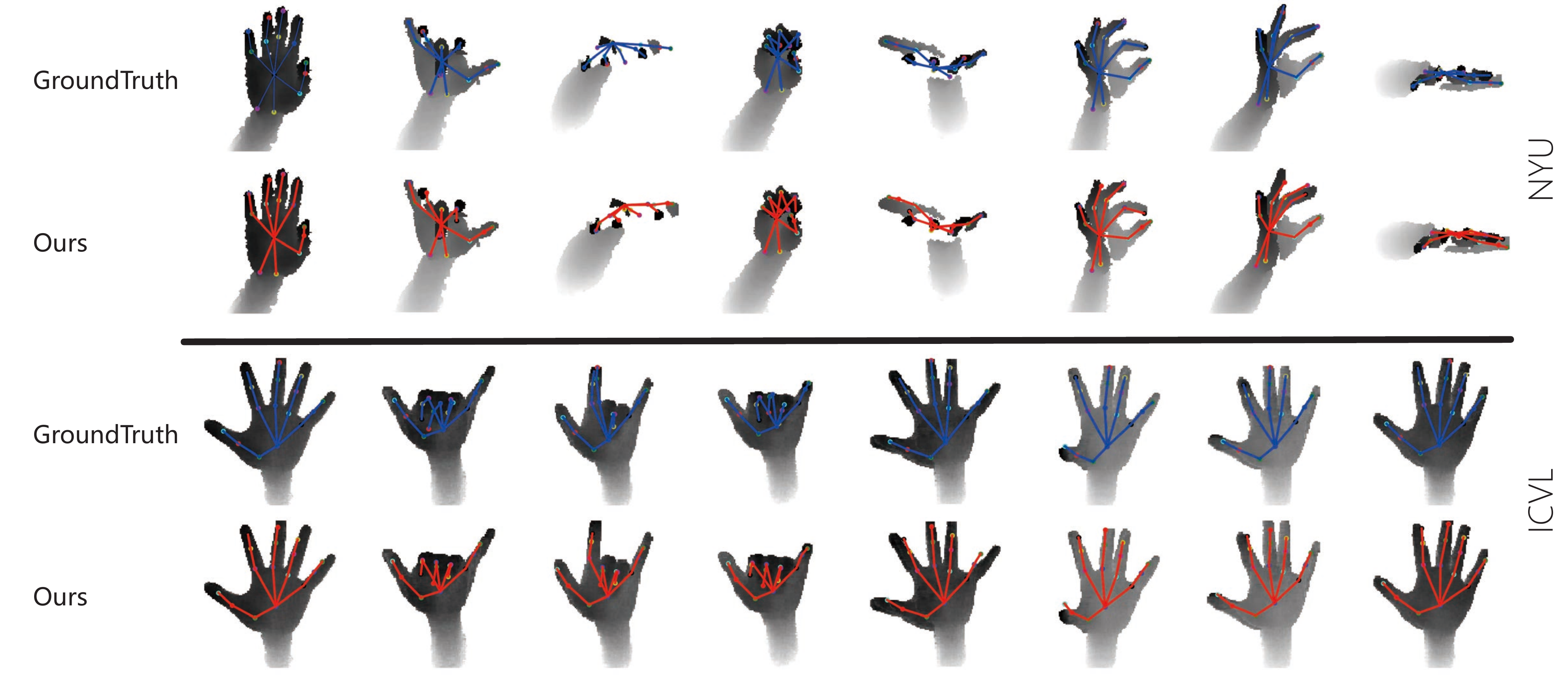}
    \caption{Qualitative results for NYU and ICVL dataset. Best viewed in color.}
\label{fig:qualitative results}
\end{figure*}

To have an intuitive sense, we present some examples as depicted in Figure~\ref{fig:qualitative results}. Our method could get an acceptable prediction
even in the extreme circumstance.

\begin{table}[htbp]
  \centering
  \caption{Mean error distance of different methods on NYU dataset.}
    \begin{tabular}{|l|c|}
    \hline
    Methods & \multicolumn{1}{l|}{Average joint error (mm)} \\
    \hline
    HeatMap~\cite{tompson2014real} & 21.02 \\
    DeepPrior~\cite{oberweger2015hands} & 19.73 \\
    Feedback~\cite{oberweger2015training} & 15.97 \\
    DeepModel~\cite{zhou2016model} & 16.90 \\
    Lie-X~\cite{xu2017lie} & \textbf{14.51} \\
    CADSTN  & 14.83 \\
    \hline
    \end{tabular}%
  \label{tab:mean error NYU}%
\end{table}%

\textbf{ICVL Dataset} contains 22k training images (totally 330k by augmentation), and is separated into ten sequences. And testing dataset
includes two sequences, there are 1.6k images in total.

On ICVL dataset, we compare our proposed method against five approaches: CrossingNet~\cite{Wan_2017_CVPR}, LRF~\cite{tang2014latent},
DeepPrior~\cite{oberweger2015hands}, DeepModel~\cite{zhou2016model} and LSN~\cite{wan2016direction}. The quantitative results
are shown in Figure~\ref{fig:comparison ICVL}. Our method is better than LRF, DeepPrior~\footnote{DeepPrior~\cite{oberweger2015hands}
only provides test results on the first sequence on ICVL, the accuracy curve is plotted based on the provided result.}, DeepModel and
CrossingNet, and is roughly same as LSN. CrossingNet employs GAN and VAE for data augmentation, and our method surpasses it when the
threshold is bigger than 9mm. Compared to the hierarchical regression framework LSN, our method is not as accurate from 14-30mm
but outperforms when the threshold is bigger than 30mm. Furthermore, Figure~\ref{fig:joint mean ICVL} reveals the same case that
fingertips estimation is generally worse than the palm of the hand. As summarized in Table~\ref{tab:mean error ICVL}, the mean
error distance of our method is the lowest in four methods, which obtains a 1.4mm error decrease than DeepPrior and is 0.16mm smaller than LSN.
\begin{table}[htbp]
  \centering
  \caption{Mean error distance of different methods on ICVL dataset.}
    \begin{tabular}{|l|c|}\hline
    Methods & \multicolumn{1}{l|}{Average joint error (mm)} \\\hline
    LRF~\cite{tang2014latent}   &  12.56\\
    DeepPrior~\cite{oberweger2015hands} &  9.42\\
    DeepModel~\cite{zhou2016model} &  11.55\\
    CrossingNet~\cite{Wan_2017_CVPR} &10.22\\
    LSN~\cite{wan2016direction} & 8.20\\
    \textbf{CADSTN}  &\textbf{8.04}\\\hline
    \end{tabular}%
  \label{tab:mean error ICVL}%
\end{table}%
\subsubsection{Ablation Study}\label{sec:ablation}
For the sake of analyzing the effects of different parts in our method, we perform an extensive ablation experiments on NYU dataset.

\begin{figure}[htbp]
\centering
    \includegraphics[width=0.6\linewidth]{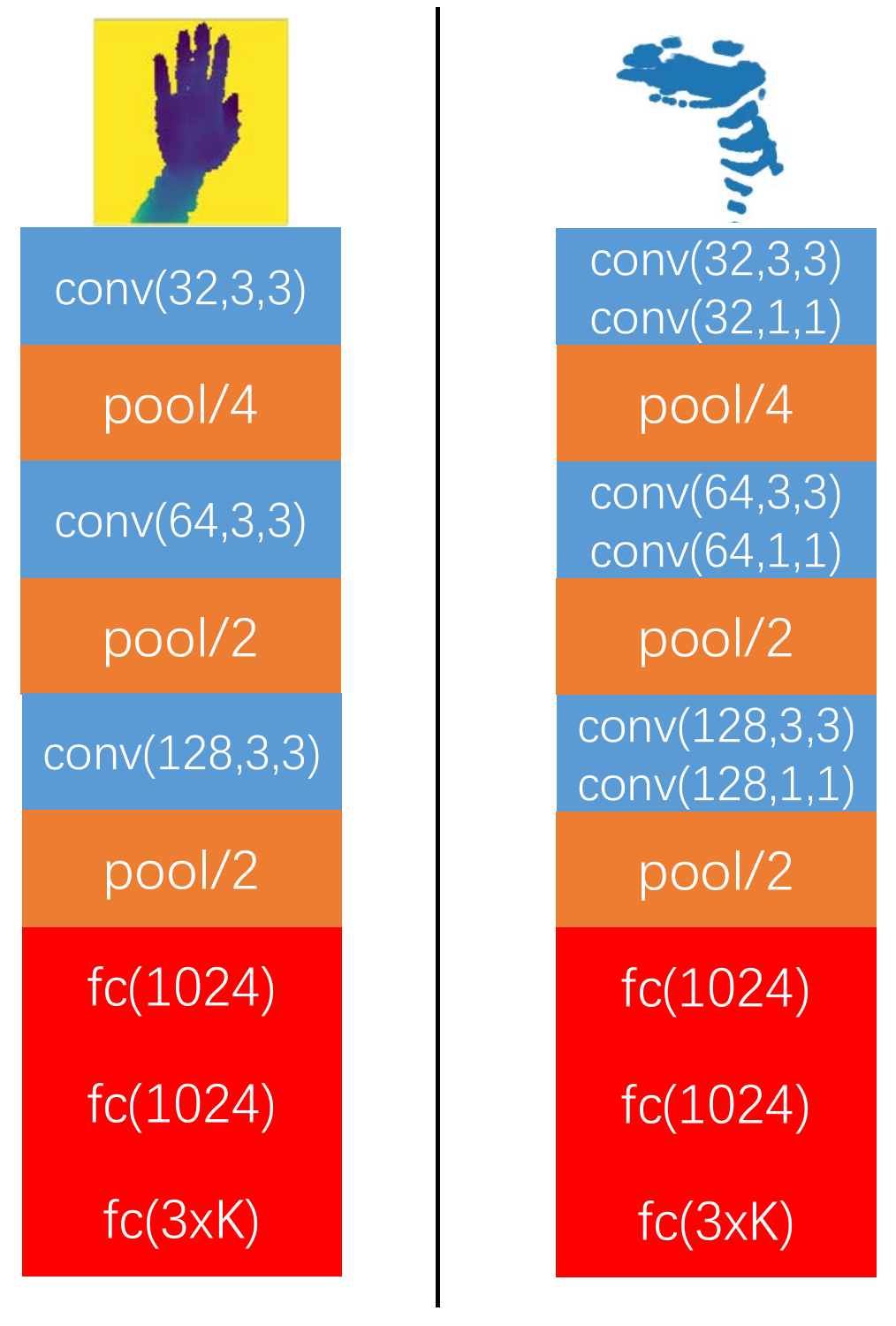}
    \caption{\emph{Baseline Regression Network} and \emph{Sliced 3D Input Network},
    these two networks are parts of \emph{Spatial Network}.}
\label{fig:baseline}
\end{figure}

\textbf{Ablation Study For Feature Representation}
Except for \emph{Spatial Network} and \emph{Temporal Network}, we train two networks named
{Baseline Regression Network} and \emph{Sliced 3D Input Network}. As shown in Figure~\ref{fig:baseline},
these two networks are parts of \emph{Spatial Network}.

\begin{figure}[htbp]\footnotesize
\centering
    \subfloat[]{
        \label{fig:ablation NYU}
        \begin{minipage}[t]{240pt}
            \raggedleft
            \includegraphics[width=1\linewidth]{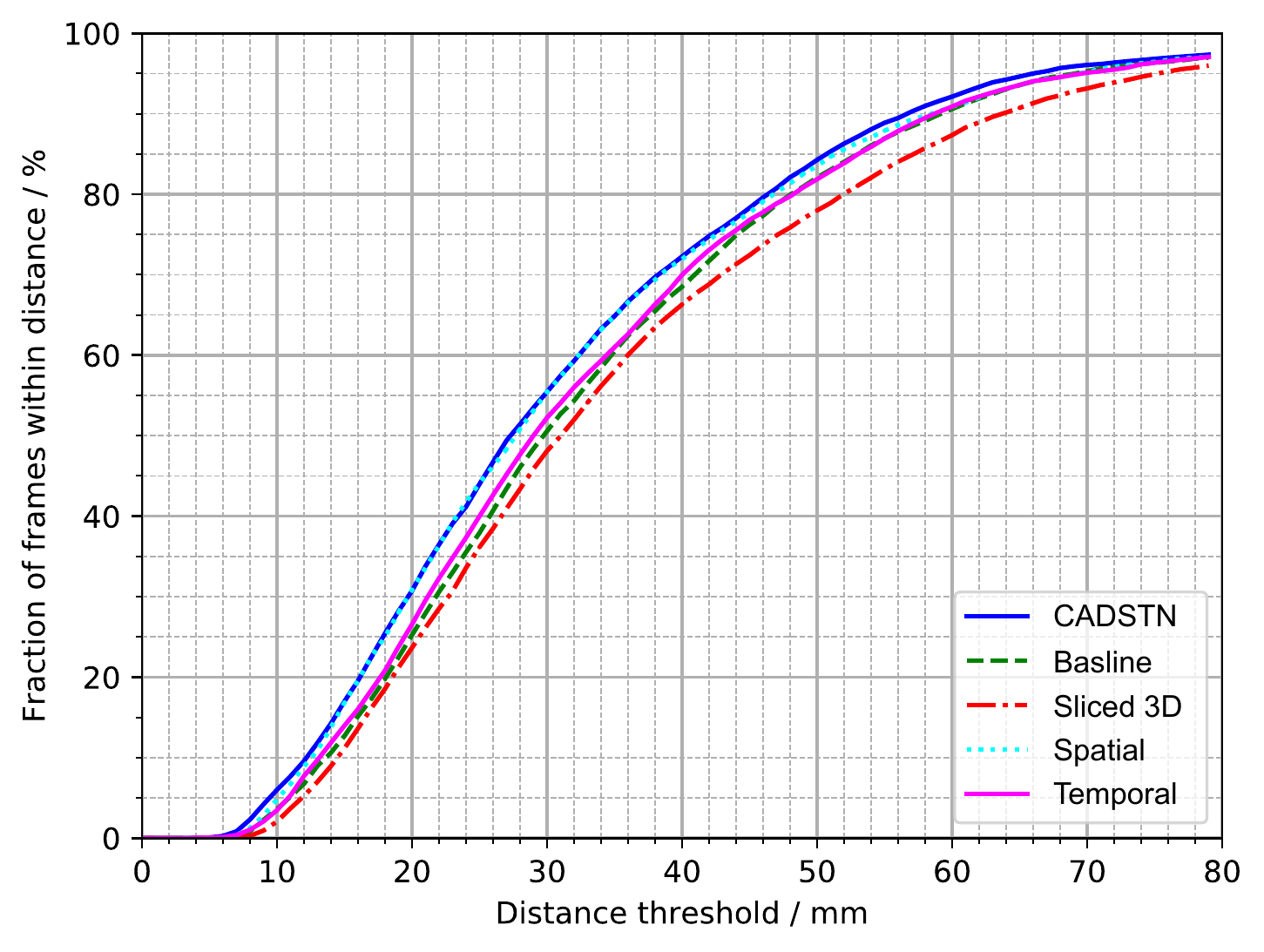}
        \end{minipage}
    }
    \\
    \subfloat[]{
        \label{fig:ablation joint mean NYU}
        \begin{minipage}[t]{240pt}
             \raggedleft
            \includegraphics[width=1\linewidth]{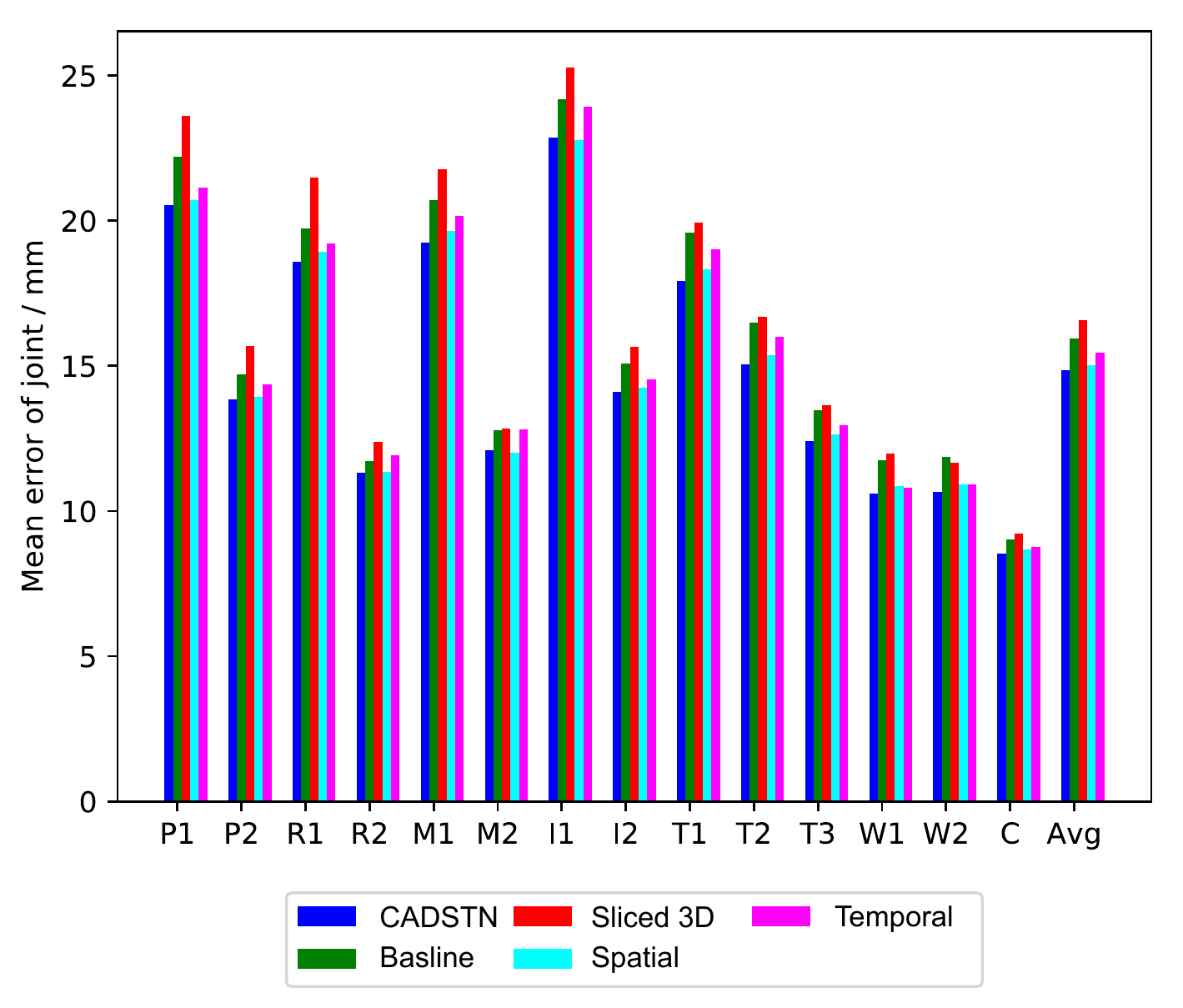}
        \end{minipage}
    }
    \caption{Performance of different part in our proposed network. Figure(a): proportion of good frames over different threshold for different part in our  proposed network (\textbf{Sliced 3D Input} is the sliced 3D volumetric representation input branch in \emph{Spatial Network} Figure(b): per-joint error distance for different parts. The palm and fingers are indexed as C: palm, T: thumb, I: index, M: middle, R: ring, P: pinky, W: wrist.}
    \label{fig:result ablation}
\end{figure}

We show the results in Figure~\ref{fig:result ablation} and Table~\ref{tab:ablation NYU}. The experimental results reveal that \emph{Spatial Network} and
\emph{Temporal Network} improve the performance by exploiting the spatial and temporal context respectively, and our final model
performs best by utilizing spatio-temporal context simultaneously.
\hspace{-3em}
\begin{table}[htbp]
  \centering
  \caption{Ablation study on NYU dataset.}
    \begin{tabular}{|l|c|}
    \hline
    Methods & \multicolumn{1}{l|}{Average joint error (mm)} \\
    \hline
    Sliced 3D Input & 16.56 \\
    Baseline Regression & 15.95 \\
    Temporal & 15.47 \\
    Spatial & 15.03 \\
    \textbf{CADSTN} & \textbf{14.83} \\
    \hline
    \end{tabular}%
  \label{tab:ablation NYU}%
\end{table}%

\hspace{-3em}

\begin{table*}[t]
\centering
\caption{Simple combination for the results from Spatial Network and Temporal Network on NYU dataset.}
\begin{tabular}{c|c|c|c|c|c|c|c|c|c|c}
    \hline
    w                          &0.1    &0.2 &0.3 &0.4 &0.5 &0.6 &0.7 &0.8 &0.9 &ours\\
    \hline
    Average joint error (mm)   &14.98  &14.94   &14.89   &14.88   &14.89   &14.90   &14.92   &15.06   &15.24   &\textbf{14.83}\\
    \hline
\end{tabular}
\label{tab:ablation_fusion}
\end{table*}
\textbf{Baseline Regression Network} is a simple network composed of convolution, max-pooling, and fully
connected layers. The network regresses the labels directly, and the average error distance is shown in Table~\ref{tab:ablation NYU}. The results
reveal that \emph{Baseline Regression Network} already surpasses some methods.

\textbf{Sliced 3D Input Network} regresses the hand joints with the sliced 3D volumetric representation. Due to the absence of
depth information, it is difficult to estimate the depth for hand joint.

\textbf{Spatial Network} hierarchically fuses the features from depth image input and
sliced 3D volumetric representation. We could find that \textbf{Sliced 3D} is the worst in several competitors. In our
opinion, the bad performance is because the depth information is sacrificed in sliced 3D volumetric representation.
Via the Deeply-Fusion network, \emph{Spatial Network} borrow the spatial information from \textbf{Sliced 3D} branch, and
it achieves a better result than \emph{Baseline Regression Network} and \emph{Sliced 3D Input Network}.

\textbf{Temporal Network} replaces the second fully connected layer with LSTM layer and
concatenates the hidden output with the input features. We find that \emph{Temporal Network} slightly improves the performance
due to the temporal information from experimental results, and the \emph{T} value for LSTM is set as 16 in training.

\textbf{CADSTN} is our proposed method, which integrates \emph{Spatial Network} and \emph{Temporal Network}
by \emph{Fusion Network}. \emph{Fusion Network} fuses the predictions from two networks and yields the final prediction.
The three networks are connected with the implicitly adaptation weights, and influence each other in the
optimization process.
Because of the temporal coherence and spatial information concerned, our method performs best in ablation experiments,
and Figure~\ref{fig:ablation joint mean NYU} reveals that the joint error distance for every joint is the lowest. Table~\ref{tab:ablation NYU}
reports that our final method decreases the average joint error by 1.12mm.

\textbf{Ablation Study For Regression}
To demonstrate the advantage of our proposed Fusion Network, we compare Fusion Network with simple combination for the results from Spatial Network and Temporal Network on NYU dataset. And the experimental results are shown in Table~\mbox{\ref{tab:ablation_fusion}}. The experimental result demonstrates our proposed Fusion Network performs better than simply combination method.

\section{Conclusion}\label{sec:conclusion}
In this paper, we propose a novel method for 3D hand pose estimation, termed \textbf{CADSTN},
which models the spatio-temporal context with three networks.
The modeling of spatial context, temporal property and fusion
are separately done by three parts.
The proposed \emph{Spatial Network} extracts depth and spatial information
hierarchically. Further on making use of the temporal coherence
between frames, \emph{Temporal Network} gives out a sequence of joints by feeding into a depth
image sequence. Then we fuse the predictions from the above two networks via \emph{Fusion Network}.
We evaluate our method on two publicly available benchmarks and the experimental results demonstrate
that our method achieves the best or the second-best result with state-of-the-art approaches and can run in real-time on two datasets.

\appendices

\section*{Acknowledgment}
This work was supported in part by NSFC under Grant U1509206 and Grant 61472353, in part by the Key Program of Zhejiang Province under Grant 2015C01027, in part by the National Basic Research Program of China under Grant 2015CB352302, and in part by the Alibaba-Zhejiang University Joint Institute of Frontier Technologies.


\ifCLASSOPTIONcaptionsoff
  \newpage
\fi

\bibliography{wu}
\bibliographystyle{IEEEtran}

\ifCLASSOPTIONcaptionsoff
  \newpage
\fi

\begin{IEEEbiography}[{\includegraphics[width=1in,height=1.25in,clip,keepaspectratio]{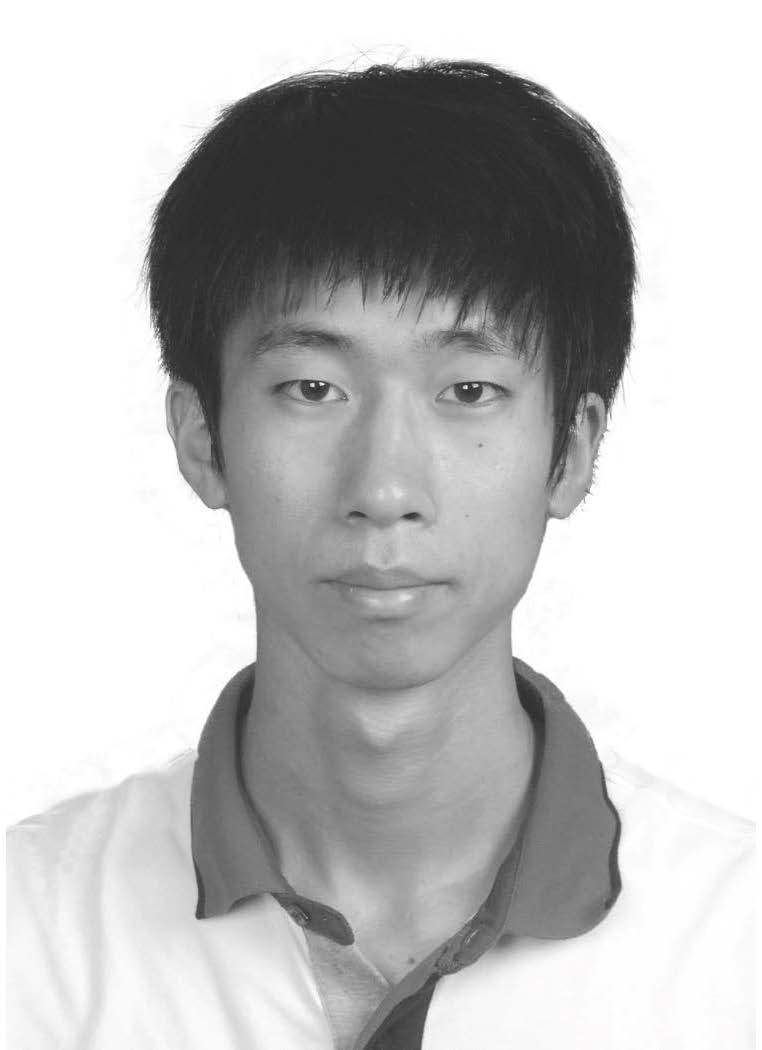}}]{Yiming Wu} is now a PhD student in College of Computer Science and Technology at ZheJiang University, Hang Zhou, China. His mentor is Professor Li Xi. Prior to that, he received a bachelor's degree in engineering from Beijing Jiaotong University. His current research direction is computer vision and machine learning.\end{IEEEbiography}

\begin{IEEEbiography}[{\includegraphics[width=1in,height=1.25in,clip,keepaspectratio]{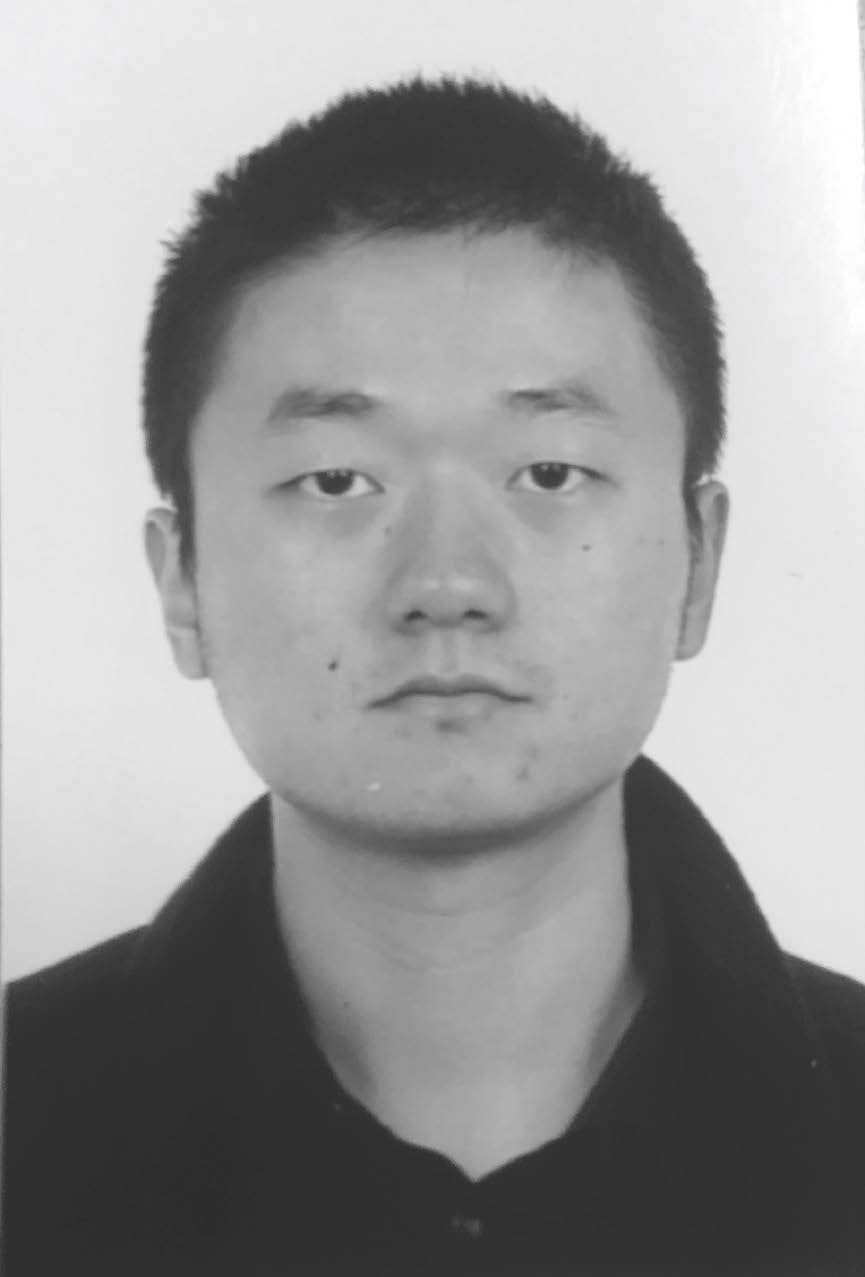}}]{Wei Ji}
is currently a third-year PhD student in College of Computer Science at Zhejiang University, Hangzhou, China. His advisors are Prof. Xi Li and Prof. Yueting Zhuang. Earlier, he received his bachelor's degree in Computer Science and Technology from Nanjing University of Science and Technology in 2015. His current research interests are primarily in computer vision and machine learning, object recognition and detection.
\end{IEEEbiography}

\begin{IEEEbiography}[{\includegraphics[width=1in,height=1.25in,clip,keepaspectratio]{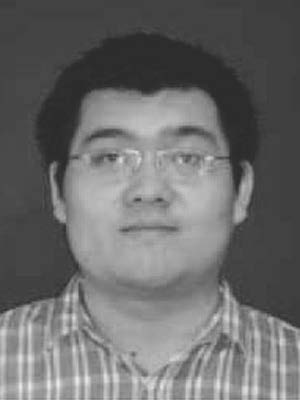}}]{Xi Li}
is currently a full professor at the Zhejiang University, China. Prior to that, he was a senior researcher at the University of Adelaide, Australia. From 2009 to 2010, he worked as a postdoctoral researcher at CNRS Telecomd ParisTech, France. In 2009, he got the doctoral degree from National Laboratory of Pattern Recognition, Chinsese Academy of Sciences, Beijing, China. His research interests include visual tracking, motion analysis, face recognition, web data mining, image and video retrieval.
\end{IEEEbiography}

\begin{IEEEbiography}[{\includegraphics[width=1in,height=1.25in,clip,keepaspectratio]{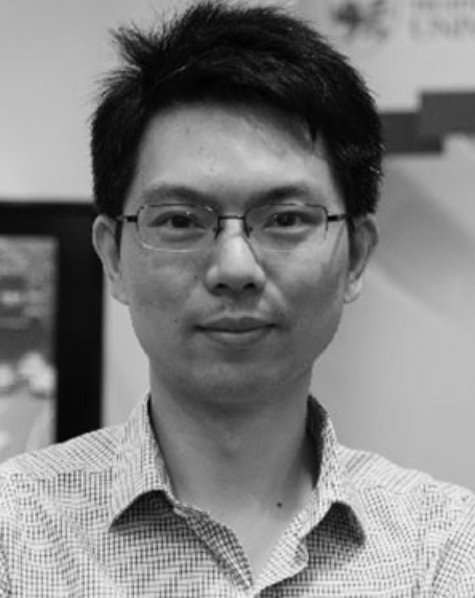}}]{Gang Wang} received the BEng degree in electrical engineering from the Harbin Institute of Technology and the PhD degree in electrical and computer engineering from the University of Illinois at Urbana-Champaign. He is an associate professor in the School of Electrical and Electronic Engineering, Nanyang Technological University (NTU), Singapore. He had a joint appointment at the Advanced Digital Science Center (operated by UIUC) as a research scientist from 2010 to 2014. His research interests include deep learning, scene parsing, object recognition, and action analysis. He is selected as a MIT Technology Review innovator under 35 for Southeast Asia, Australia, New Zealand, and Taiwan. He is also a recipient of Harriett \& Robert Perry Fellowship, CS/AI award, best paper awards from PREMIA (Pattern Recognition and Machine Intelligence Association) and top 10 percent paper awards from MMSP. He is an associate editor of the IEEE Transactions on Pattern Analysis and Machine Intelligence and an area chair of ICCV 2017. He is a member of the IEEE.
\end{IEEEbiography}

\begin{IEEEbiography}[{\includegraphics[width=1in,height=1.25in,clip,keepaspectratio]{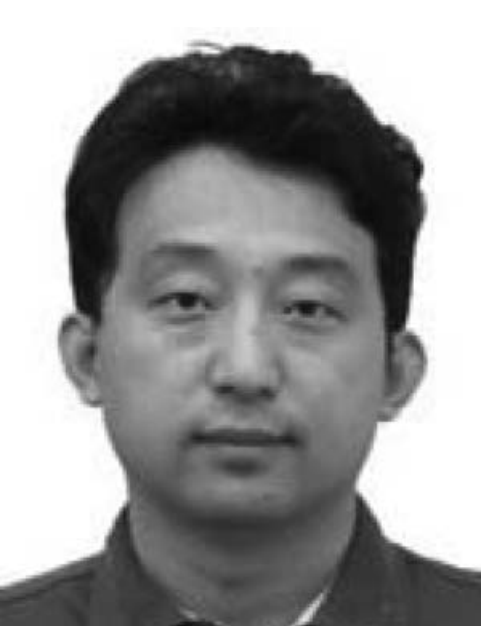}}]{Jianwei Yin} received the PhD degree in computer science from Zhejiang University, in 2001. He is currently a professor in the College of Computer Science, Zhejiang University, China. He was a visiting scholar at the Georgia Institute of Technology, Georgia, in 2008. His research interests include service computing, cloud computing, and information integration. Currently, he is the associate editor of the IEEE Transactions on Service Computing. He is a member of the IEEE.
\end{IEEEbiography}

\begin{IEEEbiography}[{\includegraphics[width=1in,height=1.25in,clip,keepaspectratio]{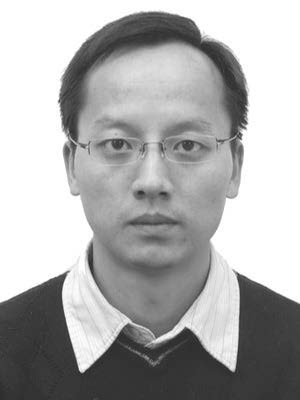}}]{Fei Wu} received the B.S. degree from Lanzhou University, Lanzhou, Gansu, China, the M.S. degree from Macao University, Taipa, Macau, and the Ph.D. degree from Zhejiang University, Hangzhou, China. He is currently a Full Professor with the College of Computer Science and Technology, Zhejiang University. He was a Visiting Scholar with Prof. B. Yu's Group, University of California, Berkeley, from 2009 to 2010. His current research interests include multimedia retrieval, sparse representation, and machine learning.
\end{IEEEbiography}
\end{document}